\documentclass{article}

\usepackage{microtype}
\usepackage{graphicx}
\usepackage{subfigure}
\usepackage{booktabs} 

\usepackage{hyperref}
\usepackage{multirow}
\usepackage{indentfirst}

\usepackage[accepted]{icml2024}
\usepackage{amsmath}
\usepackage{amssymb}
\usepackage{mathtools}
\usepackage{amsthm}
\usepackage{booktabs}
\usepackage[capitalize,noabbrev]{cleveref}

\theoremstyle{plain}

\theoremstyle{definition}

\theoremstyle{remark}

\usepackage[textsize=tiny]{todonotes}

\icmltitlerunning{A Straightforward yet Effective Approach}

\begin{document}

\twocolumn[
\icmltitle{Separating Novel Features for Logical Anomaly Detection: \\ A Straightforward yet Effective Approach}


\begin{icmlauthorlist}
\icmlauthor{Kangil Lee}{}
\icmlauthor{Geonuk Kim}{} \\ 
\vspace{2mm}

\end{icmlauthorlist}


\icmlkeywords{Machine Learning, ICML}

\vskip 0.3in
]

\begin{abstract}
Vision-based inspection algorithms have significantly contributed to quality control in industrial settings, particularly in addressing structural defects like dent and contamination which are prevalent in mass production. Extensive research efforts have led to the development of related benchmarks such as MVTec AD~\cite{bergmann2019mvtec}. However, in industrial settings, there can be instances of logical defects, where acceptable items are found in unsuitable locations or product pairs do not match as expected. Recent methods tackling logical defects effectively employ knowledge distillation to generate difference maps. Knowledge distillation~(KD) is used to learn normal data distribution in unsupervised manner.

Despite their effectiveness, these methods often overlook the potential false negatives. Excessive similarity between the teacher network and student network can hinder the generation of a suitable difference map for logical anomaly detection. This technical report provides insights on handling potential false negatives by utilizing a simple constraint in KD-based logical anomaly detection methods. We select EfficientAD~\cite{batzner2023efficientad} as a state-of-the-art baseline and apply a margin-based constraint to its unsupervised learning scheme. Applying this constraint, we can improve the AUROC for MVTec LOCO AD by 1.3~\%.
\end{abstract}
\section{Introduction}
  \begin{figure}[t]
    \begin{center}
    \centerline{\includegraphics[width=0.9\columnwidth]{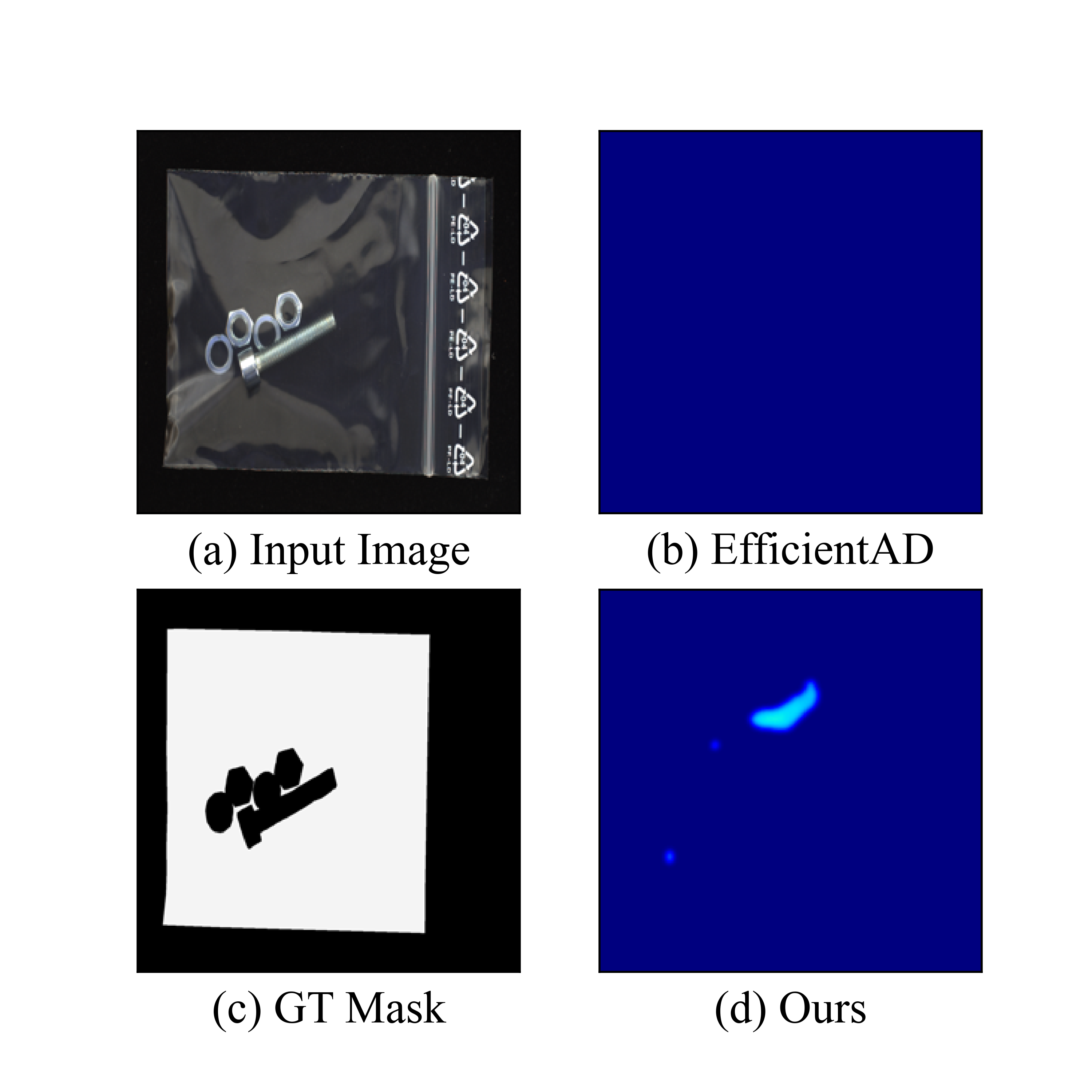}}
    \caption{It illustrates the potential false negative problem that knowledge distillation-based methods such as EfficientAD~\cite{batzner2023efficientad} can experience on some samples. (c) indicates that a bolt must be somewhere in the white area. If there is no appropriate constraint during unsupervised learning, the difference map cannot represent meaningful difference. Accordingly, given an image which has permissible objects containing logically abnormal patterns such as (a), no abnormal pattern may be capture as (b). On the other hand, when applying the proposed constraint, the existence of an logically abnormal pattern can be detected to some extent as (d).}
    \end{center}
  \end{figure}
  From a few decades ago, vision-based inspection algorithm plays an important role in the mass production of many industrial area. The algorithm usually detects anomaly patterns that should not exist in a normal product. Anomaly can be categorized in two types, structural and logical anomalies. Intuitive defect, namely structural anomaly, has been studied well by many researchers~\cite{cohen2020sub,roth2022towards,liu2023simplenet} by using related datasets such as MVTec AD~\cite{bergmann2019mvtec}. Unlike structural anomaly, logical anomaly has been garnering interest only recently. Logical anomaly usually means that a permissible object being present in an invalid location or lying in invalid order~\cite{bergmann2022beyond}. To solve the logical anomaly detection problem, it is important to extract global context~\cite{bergmann2022beyond}. Regarding the extraction of global context, the structural anomaly-oriented methods inevitably show weakness for capturing abnormal features related to the logical constraints due to the lack of ability to extract global context. Since these methods are specialized to capture local features, when if a permissible object appear in abnormal area, those cannot recognize the appearance as abnormal pattern. Hence, researchers have been digging into pulling out proper global features for logical anomaly detection~\cite{bergmann2022beyond,yao2023learning,guo2023template,zhang2024contextual}. Most prevalent approach is utilizing bottleneck paradigm to obtain the globally consistent features. For example, Bergmann et al. use two branches which are local branch and global branch. Each branch has two neural networks which are asynchronous in terms of capacity. By using capacity difference between the networks in each branch, global and local feature maps are extracted. In global branch, to capture globally abnormal pattern, Bergmann et al. intently design a small capacity network to have bottleneck for globally consistent feature extraction~\cite{bergmann2022beyond}. \\  
  In industrial settings, addressing not just logical anomaly but also structural anomaly is essential. Thus, finding an effective solution for both types of problems becomes paramount. With this context, it is natural that clearly extracting local and global features is critical. As similar as GCAD~\cite{bergmann2022beyond}, recently proposed methods construct their framework with two branches in terms of architectural perspective~\cite{bergmann2022beyond,yao2023learning,zhang2024contextual}. These methods depend on knowledge distillation to transfer generalized feature to two student networks from a single teacher network via unsupervised manner. Thus, it is intuitive that global feature extractors have risk of producing false negatives if there does not exist any difference between two feature maps from the two student networks. \\
  In this paper, we introduce a novel logical anomaly detection method built on EfficientAD~\cite{batzner2023efficientad}. We enforce the feature extractors which are used in global feature extraction to keep difference between each others. We use a teacher network for distilling knowledge to student and auto-encoder which are used to extract local and global features, respectively. Then, our constraint forces that the auto-encoder do not learn a very novel feature from the teacher network. It induces global branch keep proper difference for logical anomaly detection. With our approaches, we can achieve 92.0 \% AUROC on MVTec LOCO AD~\cite{bergmann2022beyond}. 
\section{Related Works}
\textbf{Anomaly detection for industrial images.}~In industrial environment, detecting defect in a product is very important for quality management. For a long time, vision-based anomaly detection system has been contributing hugely in this respect. Since the advent of using a pretrained deep neural network features, this field has developed remarkably~\cite{cohen2020sub,roth2022towards,bae2023pni}. 
Predominantly, the prevailing anomaly detection methods have primarily concentrated on the identification of structural anomalies. To recognize this type of anomaly, it is critical to capture local feature because most structural anomalies have small patterns. Hence, patch-memory based methods have effectively handled the structural anomalies~\cite{defard2021padim,roth2022towards,bae2023pni}. For instance, PatchCore memorize core patches of normal images in a memory bank~\cite{roth2022towards}. Then, the patches are used to detect a novel patch within a given image by comparing the patches of a given image with the memorized patches. These methods work well for structural anomaly detection within a hypothesis that permissible objects can be everywhere. \\
However, there exists limitation for logical anomalies because logically abnormal patterns are defined with permissible objects which have illegal order or location. For this rising issue, Bergmann et al. released a logical anomaly benchmark~(MVTec LOCO)~\cite{bergmann2022beyond}. Recently proposed methods address logical anomalies by architecturally partitioning feature extractors into global and local context extractors, respectively~\cite{bergmann2022beyond,guo2023template,batzner2023efficientad}.\\
In this paper, our focus is how we alleviate the false negative problem in knowledge distillation-based logical anomaly detection methods, not through architectural approaches but from a training constraint perspective. \vspace{1mm}\\
\textbf{Logical anomaly detection.}~Since logical anomaly problem is a way of coming to surface, architecturally decoupling global and local representations show promising results~\cite{yao2023learning,bergmann2022beyond,batzner2023efficientad}. Unlike structural anomaly-oriented methods, logical anomaly detection methods design their frameworks for well extracting global context because it is critical to use global context for logical anomaly detection. \\
Currently, kim et al. propose a method that uses a few-shot segmentation network to extract fine-grained global context for logical anomaly detection~\cite{kim2023few}. It hugely improves the performance of logical anomaly detection. Unfortunately, it requires representative few shot segmentation labels. Selecting representative samples takes unpredictable time. Furthermore, the patch memory used in this method must be saved at memory bank for guiding. The memory bank occupies huge memory and searching time. These limitations delay applying a vision-based inspection system. Therefore, it is hard to apply the algorithm in industrial environments where need real-time operation. \\
In this paper, we propose a novel logical anomaly detection method which can operate in real-time. Our method doesn't require costly labels for supervised learning.

\section{Methods}
In this section, we revisit EfficientAD as our baseline method. We also highlight potential issue arising from EfficientAD's unsupervised learning and introduce corresponding solution in Section \ref{our_method}. Before explaining, we briefly introduce the framework and structural differences between EfficientAD and ours. \\
EfficientAD intentionally applied augmentation to prevent the auto-encoder from learning local features. This is because Batzner et al. wanted the auto-encoder to only be involved in global feature extraction. However, the feature which EfficientAD tries to ignore may be meaningful for logical anomaly detection. Therefore, we excluded data augmentation which could be scenario dependent. In EfficientAD, the auto-encoder does reconstruct patchified features. This task can be a difficult task for an auto-encoder configured as a shallow network. Therefore, we added instance normalization layers~\cite{ulyanov2016instance} within the auto-encoder inspired by the fact that instance normalization was effective in the style transfer task.  
\subsection{Review of EfficientAD} \label{sec:review_effiad}
EfficientAD~\cite{batzner2023efficientad} extracts global and local features by using two feature maps from triplet networks named by teacher, student, and auto-encoder. To generate global and local representations, EfficientAD trains student and auto-encoder by using knowledge distillation as unsupervised learning for normal dataset. The knowledge from teacher is transferred by minimizing distances between features. The student has two output branches for global and local features corresponding to auto-encoder and teacher, respectively. Each branch is used to generate global and local representations for a given image $x$. 
\begin{equation} \label{eq1}
    \begin{split}
        d~^{SA}(x) &= ||S^{A}(x) - A(x)||_{2}^{2} \\
        d~^{TA}(x) &= ||T(x) - A(x)||_{2}^{2} \\
        d~^{TS}(x) &= ||T(x) - S^{T}(x)||_{2}^{2} \\
    \end{split}
\end{equation}
$T$, $S$, and $A$ represent the teacher, student, and auto-encoder, respectively. Here, $T(x)$, $S^{A}(x)$, $S^{T}(x)$, and $A(x)$ are tensors in $R^{C \times H \times W}$, where $C$, $H$, and $W$ denote channel, height, and width, respectively. Superscript on $S$ means output branch corresponding to teacher and auto-encoder. $|| \cdot ||_{2}^{2}$ means squared L2-distance. Except $d~^{TS}(x)$, the other squared L2-distances are minimized in the unsupervised learning. In the case of $d~^{TS}(x)$, optimization targets are restricted to the features having the values which are higher than predefined quantile.
\begin{equation} \label{eq2}
    mask~^{TS}_{c,h,w} = 
    \begin{cases}
        1 & \text{$(T_{c,h,w}(x) - S^{T}_{c,h,w}(x))^{2} \geq \tau $}\\
        0 & \text{otherwise}
    \end{cases}
\end{equation}
where, $c \in [1,C]$, $h \in [1,H]$, and $ w \in [1,W]$. The value of $\tau$ is obtained from the squared difference map between $T(x)$ and $S^{T}(x)$, corresponding to the predefined quantile. Therefore, features with values less than $\tau$ are excluded by the masks, and only the remaining features contribute to the computation of the loss function $d^{TS}(x)$. We denote this masked loss function as $d_{mask}^{TS}(x)$. The anomaly score is calculated via combination of mean difference maps between $T$ and $S$. \\
Batzner et al. normalize the difference maps by using two quantile values as similar as min-max normalization. However, one of quantile value actually is not the minimum in each difference map, the difference map can have negative values. Even if a region of one difference map represents positive score, a region of the other difference map represents negative values. Then, if we combine the two difference maps like EfficientAD, we can observe undesired cancellation. To avoid this cancellation, we simply project each difference map onto the range from 0 to 1 by using sigmoid function. Since EfficientAD generates a global representation by leveraging the channel-wise mean of difference map between $S^{A}$ and $A$, when $S^{A}$ and $A$ are close as reaching an indistinguishable level, it potentially provokes a risk of false negatives. As show in Figure 1, if there is no proper constraint in the unsupervised learning to learn normal data distribution, difference map cannot capture any information for logical anomalies. Therefore, in order to address this problem, we propose distinctive feature separation constraint.
\begin{figure}[t]
\begin{center}
\centerline{\includegraphics[width=1\columnwidth]{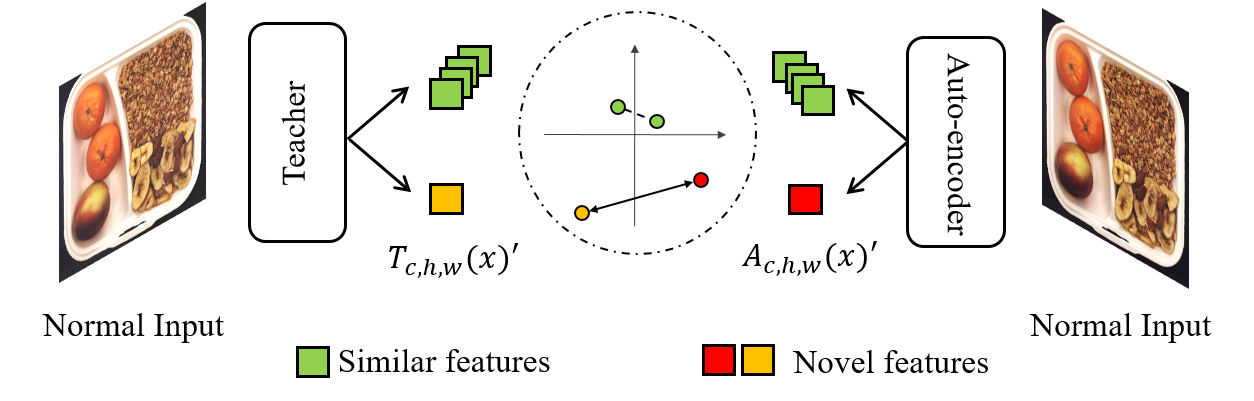}}
\caption{The overview of the proposed constraint. At training time, only normal image is fed into teacher and auto-encoder. We split output features from teacher and auto-encoder by the quantile value $\tau^{'}$ into similar and novel features. Then, we enforce auto-encoder to keep far distance by margin for novel features. Global representation is obtained from channel mean of difference map between $S^{A}$ and $A$. Since $S^{A}$ is trained on $d^{SA}(x)$ in Eq.~\ref{eq1}, when an image is given, $S^{A}$ and $A$ produce different features on novel pattern. Consequently, our constraint leads $S^{A}$ and $A$ to keep distance on the novel pattern.}
\end{center}
\end{figure}
\subsection{Distinctive Feature Separation Constraint} \label{our_method}
As outlined in section \ref{sec:review_effiad}, maintaining an appropriate separation between $S^{A}$ and $A$ is crucial for deriving meaningful global representation. $S^{A}$ learns from $A$, and $A$ acquires knowledge from $T$. Thus, manipulating the flow of information from $T$ to $A$ inevitably introduces a gap between $S^{A}$ and $A$. This intuition is our ground idea. Thus, we impose a constraint on the auto-encoder, preventing it from learning highly distinctive features from teacher. We call the constraint as Distinctive Feature Separation Constraint~(DFSC). DFSC is formed according to the following procedure. First, we find the distinctive features from the difference map between $T$ and $A$. The threshold $\tau^{'}$ is determined by the predefined quantile value among the difference map. By using $\tau^{'}$, we compute $mask^{TA}_{c,h,w}$ as Eq.\ref{eq3}
\begin{equation} \label{eq3}
    mask~^{TA}_{c,h,w} = 
    \begin{cases}
        1 & \text{$(T_{c,h,w}(x) - A_{c,h,w}(x))^{2} \geq \tau^{'} $}\\
        0 & \text{otherwise}
    \end{cases}
\end{equation}
Second, we mask out the features having smaller value than the $\tau^{'}$ as Eq.\ref{eq4}. And we perform l2-normalization over channel dimensions of $T(x)$ and $A(x)$ for stable training.
\begin{equation} \label{eq4}
    \begin{split}
        T_{c,h,w}(x)^{'} &= 
        \begin{cases}
            T_{c,h,w}(x) & \text{if $mask~_{c,h,w}^{TA} = 1 $} \\
            0 & \text{otherwise}
        \end{cases} \\
        A_{c,h,w}(x)^{'} &= 
        \begin{cases}
            A_{c,h,w}(x) & \text{if $mask~_{c,h,w}^{TA} = 1 $} \\
            0 & \text{otherwise}
        \end{cases} \\
    \end{split}
\end{equation}
Then, we calculate the L2-norm of the difference map between $T(x)$ and $A(x)$ across the channel dimension at each location $(h,w)$ as Eq.\ref{eq5}. 
\begin{equation} \label{eq5}
    \mathcal{D}_{h,w}(T, A) = \left( \sum_{c=1}^{C} (T_{c,h,w}(x)^{'} - A_{c,h,w}(x)^{'})^{2} \right)^{\frac{1}{2}}
\end{equation}
Finally, we define the constraint with a predefined constant margin $m$ as Eq.\ref{eq6}. The margin is ranged from 0 to 2.
\begin{equation} \label{eq6}
    L_{DFSC} = \frac{1}{HW} \sum_{c=1}^{C} \sum_{h=1}^{H} max~(m - \mathcal{D}_{h,w}(T, A),~0)
\end{equation}
We incorporate our constraint into the EfficientAD loss functions. Consequently, this constraint guides the auto-encoder to learn features that are relatively less discriminative. This induces $S^{A}$ and $A$ to respond sensitively when presented with a novel pattern, aiding the extraction of a global representation. The total loss function for a given image $x$ is following as Eq.\ref{eq7}.
\begin{equation} \label{eq7}
    L_{total} = d^{SA}(x) + d^{TA}(x) + d_{mask}^{TS}(x) + \alpha L_{DFSC}
\end{equation}
where alpha is a coefficient of our constraint. We always set this constant coefficient value as 2.

\section{Experiments}
\begin{table*}[t] \label{tbl: main}
\centering
\caption{Anomaly detection AUROC~(\%) scroes on MVTec LOCO AD for various anomaly detection methods. "Logical" and "Structural" in the first column mean logical and structural anomaly detection tasks. S-T and AST mean bergmann et al~\cite{bergmann2020uninformed} and Rudolph et al~\cite{rudolph2023asymmetric}, respectively. EAD-S and EAD-M mean EfficientAD small and medium models~\cite{batzner2023efficientad}.}
\vspace{1.5mm}
\resizebox{\textwidth}{!}{%
\begin{tabular}{cc|cccccccc|c}
\hline
                            & Category            & f-AnoGAN & SPADE & S-T & PatchCore  & AST & GCAD  & EAD-S  & EAD-M & Ours \\ \hline \hline
\multirow{6}{*}{\rotatebox{90}{Logical}}    & Breakfast Box       & 69.4     & 81.8 & 68.9  & 74.8 & 80.0      & 87.0 & 82.4 &  85.5    & 84.6      \\
                            & Screw Bag           & 49.7     & 46.8 & 55.5  & 57.8 & 80.1      & 56.0 & 58.1  & 56.7 &  67.9    \\
                            & Splicing Connectors & 68.8     & 73.8 & 65.4  & 79.2 & 81.8      & 89.7 & 93.0  & 95.5 &  96.9 \\
                            & Pushpins            & 59.1     & 60.5 & 59.5  & 63.6 & 65.1      & 97.5 & 98.2  & 97.7 & 99.8  \\
                            & Juice Bottle        & 82.4     & 91.9 & 82.9  & 93.9 & 91.6     & 100 & 96.7 & 98.4 &   99.9    \\ \cline{2-11} 
                            & Average             & 65.9     & 71.0 & 66.4  & 74.0 & 79.7      & 86.0 & 85.7  & 86.8  & 89.8   \\ \hline
\multirow{6}{*}{\rotatebox{90}{Structural}} & Breakfast Box       & 50.7     & 74.7 & 68.4  & 80.1 & 79.9      & 80.9 & 86.2  &  88.4   & 85.3 \\
                            & Screw Bag           & 46.1     & 59.8 & 87.0  & 92.0 & 95.9      & 70.5 & 89.4  & 90.7 &  92.4  \\
                            & Splicing Connectors & 63.8     & 57.1 & 96.8  & 88.0 & 89.4      & 78.3 & 97.9  & 98.5 &  98.1  \\
                            & Pushpins            & 74.9     & 58.1 & 90.3  & 87.9 & 77.8      & 74.9 & 97.3  & 96.1 &  95.6   \\
                            & Juice Bottle        & 77.8     & 84.9 & 99.3  & 98.5 & 95.5      & 98.9 & 99.8  & 99.7 &  99.9   \\ \cline{2-11} 
                            & Average             & 62.7     & 66.9 & 88.4  & 89.3 & 87.7      & 80.7 & 94.2  & 94.7 & 94.3     \\ \cline{2-11} 
\multicolumn{2}{c|}{Average (Total)}              & 64.3     & 68.9 & 77.4  & 81.7 & 83.7      & 83.4 & 89.9  & 90.7 & 92.0     \\ \hline
\end{tabular}%
}
\end{table*}
\textbf{Dataset.} We adopt MVTec LOCO AD~\cite{bergmann2022beyond}. The MVTec LOCO AD consists of five categories which are composed of structural and logical anomalies. It contains 3644 images which are split into training, validation, and test sets. \\
\textbf{Evaluation Metrics.} In order to evaluate our approach, we use image level AUROC~(Area Under the Receiver Operator Curve), pixel-level AUROC, and AUPRO~(Area Under Per Region Overlap). Further, we use sPRO which is a generalized version of AUPRO~\cite{bergmann2022beyond}. 
\subsection{Implementation Details}
We use batch size 1 for all experiments. Unlike EfficientAD, we did not provide different images into student and auto-encoder. We just resize images to 256 $\times$ 256 and perform normalization with ImageNet mean and standard deviation. Further, we do not use ImageNet penalty~\cite{batzner2023efficientad} for efficiency. Even if we do not include ImageNet penalty in unsupervised learning, we can achieve state-of-the-art performance. 
We add instance normalization layers on auto-encoder before all activation layers for stable training and plus relu layer on the start of decoder part. We applied the proposed method based on EfficientAD's small patch description network. 
We used AdamW as optimizer and the learning rate and weight decay are $1 \times 10^{-4}$ and $1 \times 10^{-5}$, respectively. We trained the our networks by $7 \times 10^{4}$ iterations with exponential warm up scheduler~\cite{ma2021adequacy} for all experiments. We used the $\beta_{2}$ of Ma et al. as 0.997. At 10\% remaining iterations, we reduced the learning rate by a factor of 0.1. We used exponential moving average for our student weight update~\cite{chenempirical}. The momentum coefficient is set to 0.99 for all experiments. The quantiles of $\tau$ and $\tau^{'}$ are set to 0.999 for all experiments, respectively. Since our method is built on the small version of EfficientAD denoted by EAD-S, we also use EfficientAD's anomaly score normalization method. In the normalization, two quantile values are used in EfficientAD~\cite{batzner2023efficientad}. As similar as EfficientAD, we set the quantile values for this normalization as 0.9 and 0.995, respectively. More details about the normalization can be found in the Batzner et al.
\subsection{MVTec LOCO AD}
\begin{table}[t]
\centering
\caption{Area under sPRO(\%) curve up to false positive rate per pixel of 5 \% is computed for MVTec LOCO AD. We compare our method with state-of-the art methods. EAD-S and EAD-M mean EfficientAD small and medium models~\cite{batzner2023efficientad}.}
\vspace{1.5mm}
\resizebox{\columnwidth}{!}{%
\begin{tabular}{c|cccc|c}
\hline
                    & PatchCore &  GCAD & EAD-S & EAD-M & Ours \\ \hline
Breakfast Box       & 46.0      &  50.2 & 62.1  & 64.8 & 65.7    \\
Screw Bag           & 52.2      &  55.8 & 64.5  & 66.7 & 60.0 \\
Pushpins            & 44.7      &  73.9 & 82.9  & 85.8 & 84.4 \\
Splicing Connectors & 58.6      &  79.8 & 88.1  & 89.2 & 90.6 \\
Juice Bottle        & 71.0      &  91.0 & 91.8  & 92.7 & 93.7  \\ \hline
Average             & 54.5      &  70.1 & 77.9  & 79.8 & 78.9    \\ \hline
\end{tabular}%
}
\end{table}
We measured the performance of our method for MVTec LOCO AD to assess how effective the proposed training constraint was for logical anomaly detection. We compared the measured performance with various methods such as f-AnoGAN~\cite{schlegl2019f}, SPADE~\cite{cohen2020sub}, and AST~\cite{rudolph2023asymmetric}. The evaluation metrics of MVTec LOCO AD are AUROC and sPRO. For an accurate assessment of logical anomaly detection performance, we measured the AUROC by segmenting the data by category. To evaluate our method, we utilized the evaluation protocol provided by MVTec LOCO AD benchmark~\cite{bergmann2022beyond}. \\
Table 1 shows AUROC of the proposed method and the other state-of-the-art methods. Upon comprehensive examination of Table 1, a discernible positive correlation emerges between the performance of structural anomaly detection and logical anomaly detection. Base on this observation, it is believed that EfficientAD demonstrates a competitive performance on MVTec LOCO AD by judiciously combining the advantages of patch-based and difference map-based methods. Despite the successful integration, EfficientAD overlooked the potential false negative problem. To address this, we introduced distinctive feature separation constraint that alleviate this problem, resulting in improved performance for MVTec LOCO AD compared to EfficientAD. As shown in Table 1, the proposed method consistently achieved high rankings in each category for logical anomaly detection, establishing itself as state-of-the-art on average. It also demonstrated commendable performance in the structural anomaly detection, securing a competitive position. When considering both aspects, the proposed method recorded a 1.2\% higher AUROC compared to the current state-of-the-art. This result highlights its overall superiority. \\
Looking at Table 1 by category, we can see that AST shows remarkably good performance for the logical anomaly detection of Screw Bag category. This could be attributed to the utilization of positional encoding during the training. This is because the logical anomaly data in the Screw Bag category is comprised of whether or not appropriate paring is in place. We consider the combination of positional encoding and features contributes to the competitive performance for logical anomaly detection related to pairing, as it allows AST to capture a contextual understanding similar to text. However, with respect to Pushpins category, AST shows relatively low performance. We consider that it is also caused by the positional encoding. Hence, the decision to employ positional encoding or not should be determined by the inherent characteristics of the data.  \\
Table 2 represents the localization performance for each category. The proposed method exhibits the highest localization performance across three categories. However, for the Screw Bag category, which shows the most significant improvement in AUROC, the localization performance is relatively lower. When measuring localization performance at a 30\% false positive rate, the proposed method achieves an approximately similar sPRO value compared to EAD-M. Based on this observation, we infer that intentionally keeping relatively large number of locations be distinct results in more false positives in terms of localization perspective. Nevertheless, we assert that the proposed method possesses competitive localization performance compared to the latest methods.

\begin{figure*}[t]
    \begin{center}
    \centerline{\includegraphics[width=1\textwidth]{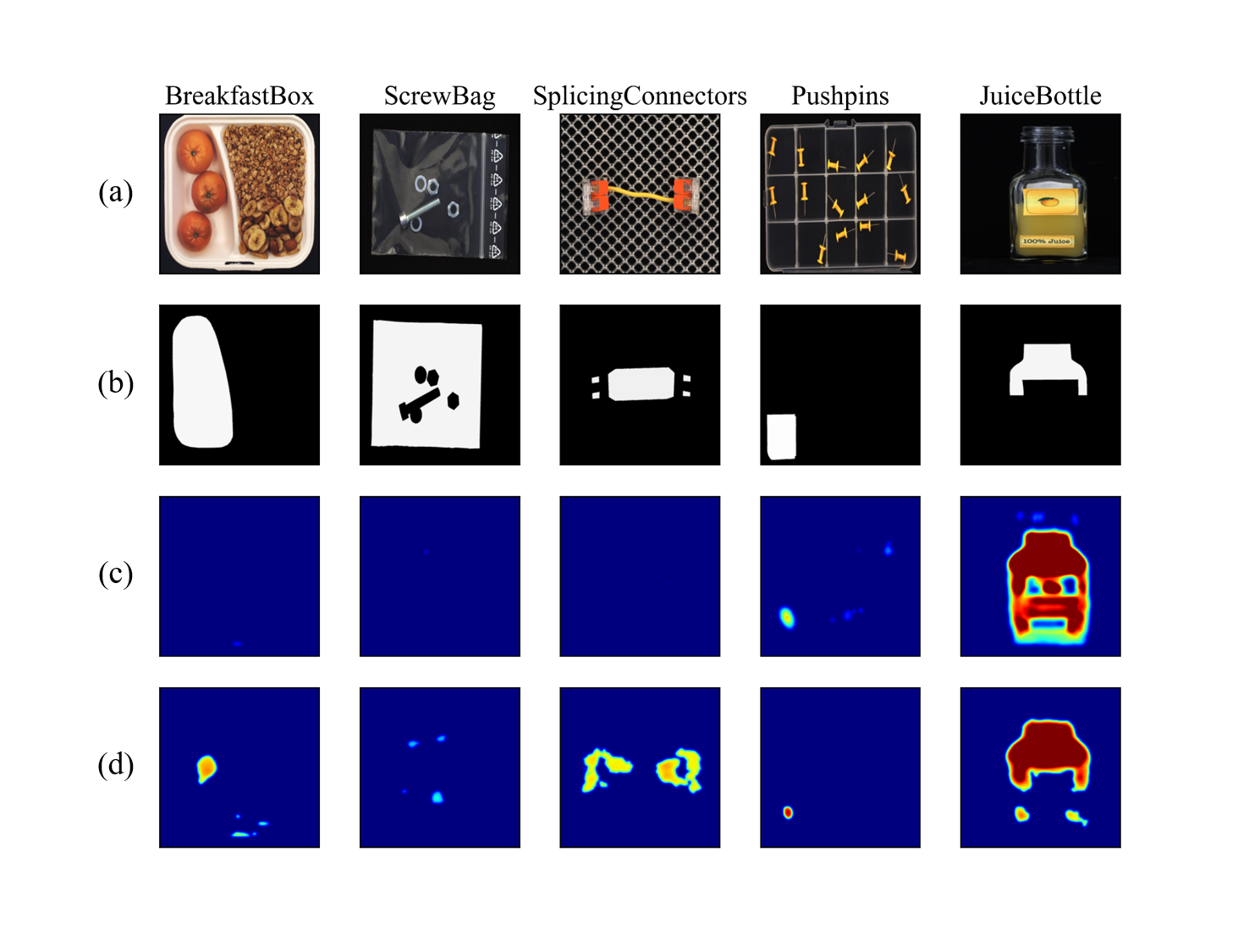}}
    \caption{(a) and (b) represent input images and ground truth masks for logical anomalies. If a model captures logical anomalies in the white areas of (b), those are considered as correct. (c) is the predicted anomaly maps produced by EfficientAD-S denoted as EAD-S. (d) represents the predicted anomaly maps produced by our method. }
    \end{center}
\end{figure*}

\begin{figure}[t]
    \begin{center}
        \centerline{\includegraphics[width=1\columnwidth]{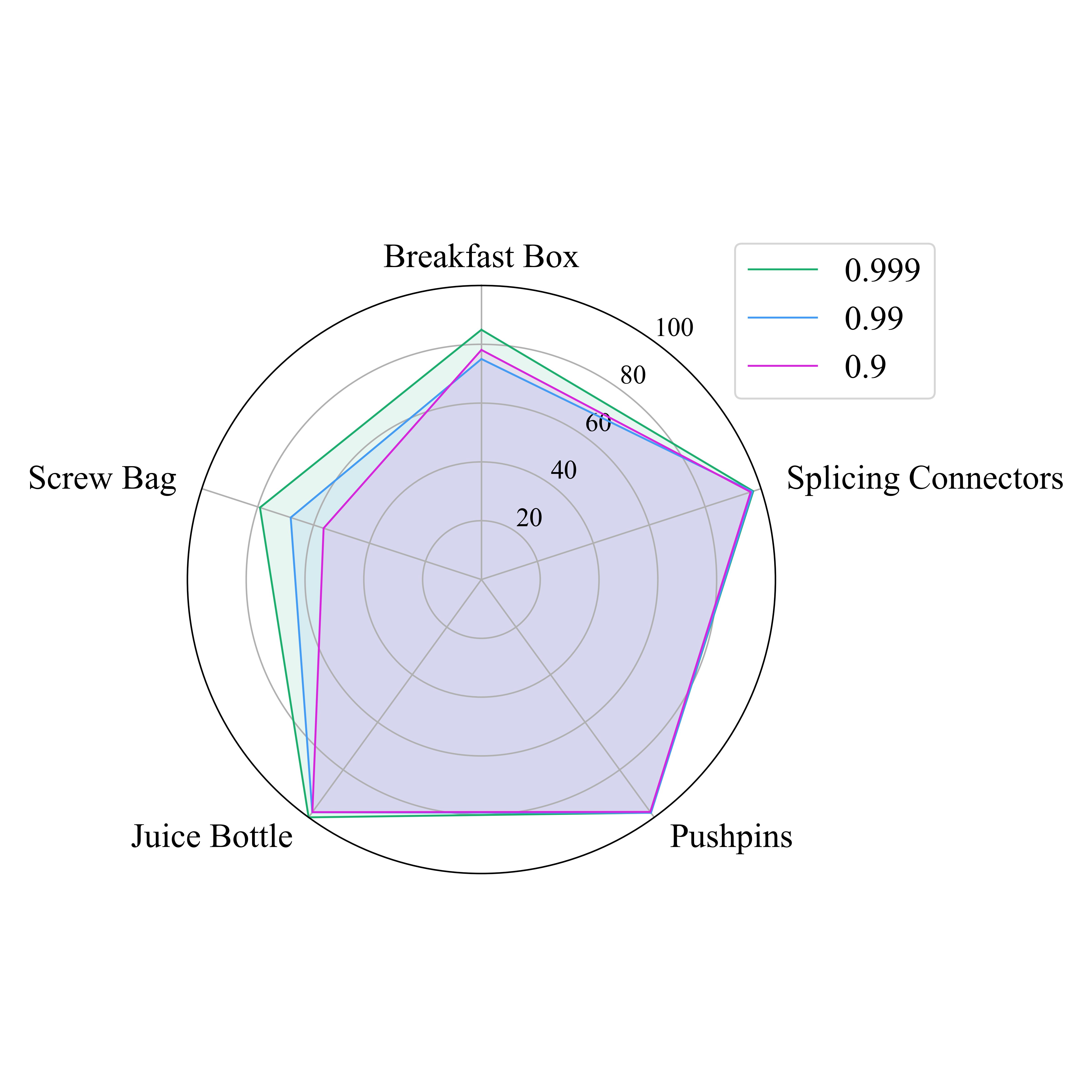}}
        \caption{Performance variations versus quantile values for MVTec LOCO AD. Evaluation metric is image-level AUROC~(\%). The legends mean the quantile values for our constraint.}
    \end{center}
\end{figure}
\subsection{Analysis}
\begin{table}[t]
\caption{Performance variation according to margin values. Evaluation metric is AUROC~(\%). The first row represents the margin value.}
\vspace{1.5mm}
\resizebox{\columnwidth}{!}{%
\begin{tabular}{c|ccccc}
\hline
                    & 0.0   & 0.2   & 0.4   & 1.0   & 2.0   \\ \hline
Breakfast Box       & 81.77 & 82.60 & 83.17 & \textbf{84.95} & 82.15 \\
Screw Bag           & 68.93 & 68.20 & 71.85 & 75.3  & \textbf{79.20} \\
Pushpins            & 96.96 & 97.34 & \textbf{97.70} & 95.77 & 93.38 \\
Splicing Connectors & 96.30 & 96.85 & \textbf{97.50} & 96.00 & 88.56 \\
Juice Bottle        & 99.77 & 99.83 & \textbf{99.99} & 99.81 & 99.76 \\ \hline
\end{tabular}%
}
\end{table}
\begin{table}[t] \label{exp:ablation}
\caption{Ablation Study on MVTec LOCO AD. AUROC~(\%) is used to measure effectiveness of each approach. The AUROC is the average AUROC over MVTec LOCO AD. The first row refers to the performance of vanilla EfficientAD when no approach is applied. Note that the augmented image is not given to the auto-encoder, and also the ImageNet penalty~\cite{batzner2023efficientad} is not used.}
\vspace{1.5mm}
\resizebox{\columnwidth}{!}{%
\begin{tabular}{cccc|c}
\hline
IN+ReLU & Sigmoid & DFSC & Momentum & AUROC \\ \hline \hline
        &         &      &          &  87.92  \\
  \checkmark      &         &      &        &  89.11     \\
  \checkmark      &  \checkmark       &      &         &  88.94 \\
  \checkmark      &  \checkmark      & \checkmark  &    & 91.22     \\
  \checkmark      &  \checkmark       & \checkmark     &   \checkmark       &    92.00   \\ \hline
\end{tabular}%
}
\end{table}
In this section, we analyze what is important factor for our approach. First, we explore the influence of margin for our constraint. Second, we investigate performance variation depending on the quantile value. Third, we compare performance changes based on ablation study. Lastly, we assess the validity of our intuition by examining qualitative results.\\
The margin parameter specifies the minimum distance that has to be kept between novel features by the teacher and the auto-encoder. Hence, the margin can be viewed as the quantitative value standing for impact of our proposed constraint on the training. Table 3 shows performance changes corresponding to variations in margin values. As depicted in Table 3, optimal margin value is different for each category. This phenomenon can be attributed to the inherent differences in data distribution across categories. Nevertheless, what remains certain is the performance improvement upon the applying the proposed constraint with a proper margin. Specifically, the screw bag category exhibited the most substantial performance improvement in terms of AUROC for logical anomaly detection. This category, characterized by the highest number of false negatives for EfficientAD, experienced the most pronounced effectiveness of the proposed constraint. This result implies that our intuition is valid. \\
According to the predefined quantile value, the features of the teacher and auto-encoder are filtered by $mask^{TA}_{c,h,w}$. Since the quantile value has a great influence on the proposed constraint, we explored performance variation produced by it. Figure 4 shows the result of our exploration over quantile values. As shown in Figure 4, low quantile value enforces auto-encoder to learn relatively small portion of teacher's knowledge, leading to low performance. Therefore, it is advisable to employ a quantile value in proximity to 0.999. \\
As shown in Table 5, we used EfficientAD's small model as a baseline and measured performance changes when applying various approaches. \\
We conducted an ablation study using EfficientAD's Small Patch Description Network (PDN) as a baseline, without applying our various approaches. Table 5 shows the average AUROC (\%) for five categories of MVTec LOCO AD when the proposed approaches are applied. IN + ReLU denotes the inclusion of instance normalization layers and additional ReLU within the auto-encoder. Sigmoid represents passing the anomaly score map through the sigmoid function to map it to the range of 0 to 1. DFSC indicates applying the Distinctive Feature Separation Constraint. Lastly, Momentum signifies the use of momentum update when updating the student PDN. \\
When the proposed DFSC is applied, the most substantial performance improvement is observed, with the next most influential factor being the enhancement of the auto-encoder's architecture. These findings underscore the significance of global representation, aligning with the discussion in GCAD concerning the handling of logical anomaly detection. The use of momentum update, updating the student PDN with a small proportion at each iteration, also contributed to improving performance by inducing the extraction of appropriate difference maps. While there was a slight degradation from a detection perspective when using sigmoid, there is an improvement from a localization perspective. Therefore, it is evident that each of the proposed approaches played a pivotal role in significantly improving the performance of logical anomaly detection. \\
We qualitatively confirmed the effect of the proposed constraint for each category. Figure 3 shows qualitative results to compare ours with small version of EfficientAD denoted by EAD-S. As shown in Figure 3.(c) for BreakfastBox category, EAD-S cannot capture logical anomaly in the white area of Figure 3.(b). In contrast, if we apply our constraint in the unsupervised learning, we can see that the logical anomaly is captured as Figure 3.(d). Similarly, we can observe that the proposed method can spot the logical anomalies relatively better for ScrewBag and SplicingConnectors categories. In the case of the remaining two categories, Pushpins and JuiceBottle, as can be seen in Table 2, those are qualitatively better than EAD-S in terms of the localization performance. Therefore, we can confirm qualitatively that it is effective to apply the proposed DFSC to ensure that the difference map maintains an appropriate distance.

\section{Conclusion}
In this paper, we claim that potential false negative problem is overlooked by difference map-based logical anomaly detection methods. Since most difference map-based methods use knowledge distillation for unsupervised learning to learn normal data distribution, two neural networks which produce difference map could be exposed to a risk to be close as reaching an indistinguishable level. This problem is especially noticeable in neural networks used to extract global representation, which is important for logical anomaly detection. \\
To tackle this issue, we propose distinctive feature separation constraint to keep proper distances between neural networks for logical anomaly detection with minor architectural modifications. Our constraint prevents the teacher and auto-encoder from having excessive similarity. When applying the proposed method, performance in terms of Image level AUROC is improved by 1.3\% compared to the EfficientAD-M for MVTec LOCO. Additionally, compared to the small version of EfficientAD, which we used as a baseline, the sPRO value is improved by 1 \%. The improvement of localization performance is also found by qualitative results. According to these experimental observations, it is clear that potential false negatives should be handled in the KD-based logical anomaly detection method.

\clearpage
\bibliography{biblography}
\bibliographystyle{icml2024}



\end{document}